\documentclass[conference]{IEEEtran}
\IEEEoverridecommandlockouts

\usepackage{amsmath,amssymb,amsfonts}
\usepackage{hyperref}
\usepackage{url}
\usepackage{graphicx}
\usepackage{xcolor}
\usepackage[caption=false]{subfig}
\usepackage{lipsum}

\usepackage{cite}

\usepackage[inline]{enumitem} %

\usepackage{multirow} %
\usepackage{booktabs} %
\usepackage{siunitx} %
\usepackage{etoolbox}           %
\renewcommand{\bfseries}{\fontseries{b}\selectfont} %
\robustify\bfseries             %
\newrobustcmd{\B}{\bfseries}    %
\newcommand{\itseries}{\fontshape{it}\selectfont} %
\robustify\itseries
\newrobustcmd{\IT}{\itseries} 

\usepackage{xspace}

\newbool{2d_std_as_subfig}
\booltrue{2d_std_as_subfig}

\newbool{stochasticity_x_k}
\booltrue{stochasticity_x_k}
\boolfalse{stochasticity_x_k}

\newcommand{\new}[1]{#1}

\renewcommand{\t}{\mathbf{z}}
\newcommand{\x}{{\mathbf{x}}}

\newcommand{\z}{{\mathbf{z}}}

\newcommand{\te}{\tilde{\t}}
\newcommand{\xe}{\tilde{\x}}

\newcommand{\ze}{\tilde{\z}}

\newcommand{\xh}{\hat{\x}}

\newcommand{\zh}{\hat{\z}}

\newcommand{\xtot}{\x \rightarrow \te}

\newcommand{\ttox}{\t \rightarrow \xe}

\def\AS{1.3}
\newcommand{\pixp}{pix2pix\space} %
\newcommand{\CyG}{CycleGAN\xspace} %

\newcommand{\TRBpairdOne}{${\text{TURBO}}^{\text{paired (w } \mathcal{D} \text{)}}_{\text{CNN-RESNET-CNN}}$\xspace}

\newcommand{\TRBpairdUNET}{${\text{TURBO}}^{\text{paired (w/o } \mathcal{D} \text{)}}_{\text{UNET}}$\xspace}

\newcommand{\Palette}{Palette\xspace}

\newcommand{\Let}{\mathcal{L}_{\tilde{\mathrm{z}}}(\z, \ze)}
\newcommand{\Lep}{\mathcal{L}_{\tilde{\mathrm{x}}}(\x, \xe)}
\newcommand{\Lrt}{\mathcal{L}_{\hat{\mathrm{z}}}(\z, \zh)}
\newcommand{\Lrp}{\mathcal{L}_{\hat{\mathrm{x}}}(\x, \xh)}

\newcommand{\Det}{\mathcal{D}_{\tilde{\mathrm{z}}}(\z, \ze)}
\newcommand{\Dep}{\mathcal{D}_{\tilde{\mathrm{x}}}(\x, \xe)}
\newcommand{\Drt}{\mathcal{D}_{\hat{\mathrm{z}}}(\z, \zh)}
\newcommand{\Drp}{\mathcal{D}_{\hat{\mathrm{x}}}(\x, \xh)}

\begin{document}
\bstctlcite{MyBSTcontrol} %
\renewcommand{\figureautorefname}{Fig.\negthinspace}

\title{Stochastic Digital Twin for Copy Detection Patterns
}

\author{\IEEEauthorblockN{Yury Belousov, Olga Taran, Vitaliy Kinakh and Slava Voloshynovskiy}
	\IEEEauthorblockA{\textit{Department of Computer Science, University of Geneva, Switzerland} \\
		\{yury.belousov, olga.taran, vitaliy.kinakh, svolos\}@unige.ch}
}

\maketitle

\begin{abstract}
Copy detection patterns (CDP) present an efficient technique for product protection against counterfeiting. However, the complexity of studying CDP production variability often results in time-consuming and costly procedures, limiting CDP scalability. Recent advancements in computer modelling, notably the concept of a ``digital twin'' for printing-imaging channels, allow for enhanced scalability and the optimization of authentication systems. Yet, the development of an accurate digital twin is far from trivial. 

This paper extends previous research which modelled a printing-imaging channel using a machine learning-based digital twin for CDP. This model, built upon an information-theoretic framework known as ``Turbo'', demonstrated superior performance over traditional generative models such as CycleGAN and pix2pix. However, the emerging field of Denoising Diffusion Probabilistic Models (DDPM) presents a potential advancement in generative models due to its ability to stochastically model the inherent randomness of the printing-imaging process, and its impressive performance in image-to-image translation tasks. 

This study aims at comparing the capabilities of the Turbo framework and DDPM on the same CDP datasets, with the goal of establishing the real-world benefits of DDPM models for digital twin applications in CDP security. Furthermore, the paper seeks to evaluate the generative potential of the studied models in the context of mobile phone data acquisition. Despite the increased complexity of DDPM methods when compared to traditional approaches, our study highlights their advantages and explores their potential for future applications.

\end{abstract}


\IEEEpeerreviewmaketitle

\section{Introduction}
\label{sec:introduction}

The recent upsurge in the utilization of Copy Detection Patterns (CDP), as described in \cite{picard2004digital, picard2021counterfeit, yadav2019copy, ho2014authentication}, %
has emerged as a viable method for safeguarding products against counterfeiting practices. However, the exploration of variability inherent in CDP production represents a process that is both time-intensive and financially demanding. This process necessitates the acquisition of vast volumes of data, a requirement that places a significant constraint on the scalability of the approach to incorporate new products, manufacturing technologies, and imaging devices. Consequently, the expansive adoption and continued research into CDP are impeded.

To overcome these limitations and promote the ongoing advancement of CDP, an approach involving computational modeling of the entire production pipeline, incorporating the printing and imaging channels, has been proposed. This method is referred to as a \textit{digital twin} \cite{Belousov2022wifs}. Through this approach, a comprehensive and accurate simulation of the production process is generated, allowing for an efficient examination and optimization of CDP without the traditional restrictions associated with physical data acquisition. This approach offers considerable potential for increasing the efficiency and effectiveness of anti-counterfeiting measures based on CDP.

\begin{figure}[t!]
    \includegraphics[width=1.\columnwidth]{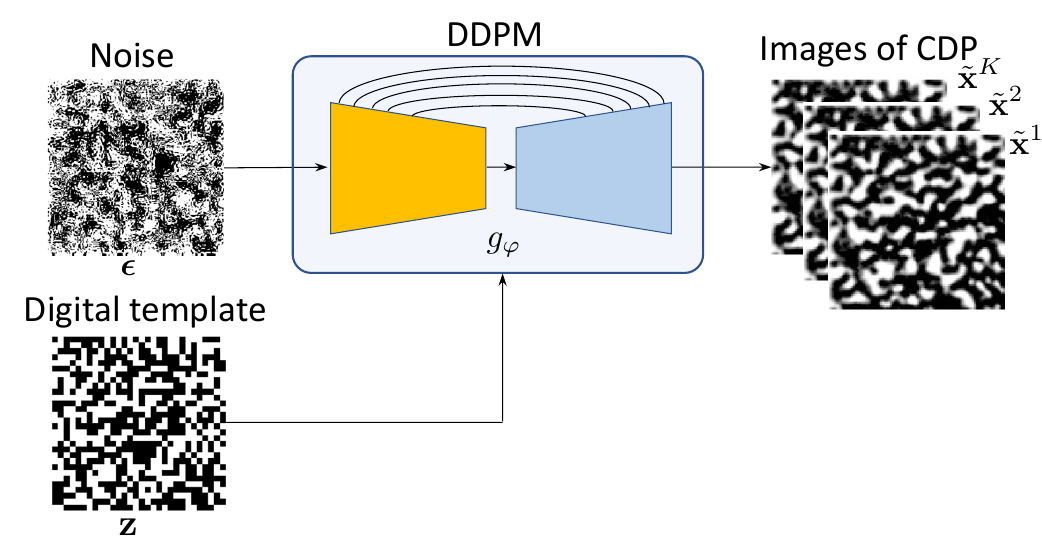}
    \caption{Schematic block-diagram of investigated DDPM generative model $g_{\varphi}$. The DDPM generative model can generate $K$ synthetic CDP $\{\xe_i^k\}_{k=1}^K$ %
    from digital template $\z$ and vice versa generate $K$ synthetic templates $\{\ze_i^k\}_{k=1}^K$ %
    for a given $\x$.  We show only the first case. The stochasticity of the generative process is ensured by different noise realizations $\boldsymbol{\epsilon}$.  }
 \label{fig:ddpm scheme}
\end{figure}

The design of \textit{digital twin} for printing-imaging channels is not a trivial task but it is crucial for both the defender and attacker. If successful, it will enable the overall optimi\new{z}ation of the whole authentication system and, in particular, the optimi\new{z}ation of the estimation of digital templates from the physical samples and synthesis of CDP images from the corresponding digital templates. Moreover, it simplifies the modeling of the intra-class variabilities and the investigation of adversarial examples.
\new{The number of training pairs needed for \textit{digital twin} is small (in the order of hundreds), while the trained model can be applied to millions of unseen digital templates.}

{
\newcommand{\sample}{000500}
\newcommand{\IMGSIZE}{0.11}

\newcommand{\genimg}[2]{\includegraphics[width=\IMGSIZE\textwidth, keepaspectratio]{images/stochasticity/#1/\sample/#2.pdf}}

\newcommand{\tximg}[1]{\genimg{t2x}{#1}}
\newcommand{\xtimg}[1]{\genimg{x2t}{#1}}
\centering

\begin{figure*}[t!]
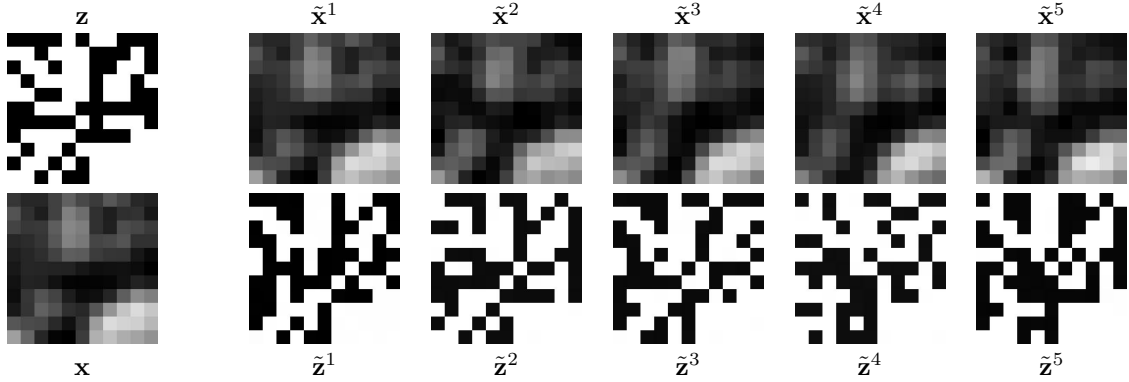

\centering
    \begin{tabular}{ c @{\hspace{35pt}} c c c c c }
    \ifbool{stochasticity_x_k}{
    $\z$ &  \multicolumn{5}{c}{$\xe^k$} \\
    }{$\z$ &  $\xe^1$ & $\xe^2$ & $\xe^3$ & $\xe^4$ & $\xe^5$ \\}    
    
    \tximg{template} & \tximg{pred_1} & \tximg{pred_2} & \tximg{pred_3} & \tximg{pred_4} & \tximg{pred_5} \\
    
    \tximg{pred_1} & \xtimg{pred_1} & \xtimg{pred_2} & \xtimg{pred_3} & \xtimg{pred_4} & \xtimg{pred_5} \\

    \ifbool{stochasticity_x_k}{
    $\x$ &  \multicolumn{5}{c}{$\ze^k$} \\
    }{$\x$ &  $\ze^1$ & $\ze^2$ & $\ze^3$ & $\ze^4$ & $\ze^5$ \\}    
    
    \end{tabular}
    \caption{The stochasticity in the DDPM Model. The first column displays the original digital template $\t$ at the top and its counterpart physical sample $\x$ at the bottom. The subsequent columns show the stochastic estimations of the CDP images $\{\xe^k\}_{k=1}^5$ and the digital templates $\{\te^k\}_{k=1}^5$, generated by the DDPM based on the \Palette framework. To enhance visual comprehension, only an enlarged $11 \times 11$ central crop is displayed.
    }
    \label{fig:stochasticity}
\end{figure*}
}
The current work is a continuation of our previous work \cite{Belousov2022wifs} that was dedicated to modeling a printing-imaging channel using a machine learning-based \textit{digital twin} for CDP. The model studied in \cite{Belousov2022wifs} is based on an information-theoretic framework called \textit{Turbo}. 
In our current work, we aim at %
comparing Turbo to the Denoising Diffusion Probabilistic Models (DDPM)\cite{ho2020denoising}, which present a popular family of modern generative models \new{(Fig. \ref{fig:ddpm scheme})}. DDPM model the prior data distribution via a diffusion process. Recently the DDPM methods demonstrated remarkable performance in the image-to-image translation tasks and outperformed many  state-of-the-art models based on GAN-like architectures \cite{dhariwal2021diffusion}. In contrast to the GAN-based generators that are mostly deterministic in nature, DDPM allows stochastic outputs, i.e., the different outputs for the same input. Taking into account the natural randomness of the printing-imaging process, the stochasticity of the synthesised twins is a key factor for high-precision simulation of real CDP. Besides this valuable advantage, the DDPM methods have high complexity compared to the traditional approaches. That is why the study of real advantages of DDPM based CDP digital twins represents a great practical interest.

In our previous work \cite{Belousov2022wifs} we demonstrated the superiority of the Turbo framework over the state-of-the-art generative models. The main goal of this study is to compare Turbo with DDPM on the same CDP datasets and to establish the real advantages of DDPM models. Moreover, we aim at evaluating the generative capabilities of the models in the context of mobile phone data acquisition.

{
\newcommand{\STDIMGSIZE}{0.17}

\captionsetup[subfigure]{labelformat=empty}

\begin{figure*}[t!]
    \centering
    \ifbool{2d_std_as_subfig}{
    \subfloat[$\z$ \label{fig:real_z_iphone}]{\includegraphics[height=\STDIMGSIZE\textwidth]{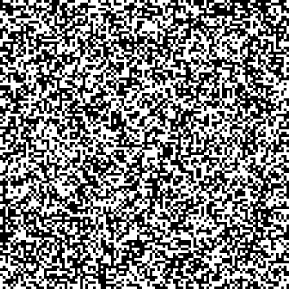}}
    \quad 
    \subfloat[iPhone $\xtot$  \label{fig:iphone x2t 2}]{\includegraphics[height=\STDIMGSIZE\textwidth]{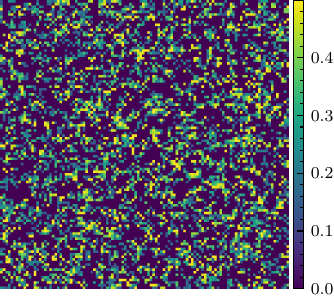}}
    \qquad \qquad
    \subfloat[iPhone $\x$ \label{fig:real_x_iphone}]{\includegraphics[height=\STDIMGSIZE\textwidth]{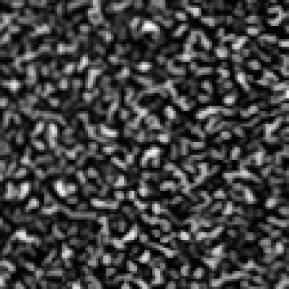}}
    \quad     
    \subfloat[iPhone $\ttox$ \label{fig:iphone t2x}]{\includegraphics[height=\STDIMGSIZE\textwidth]{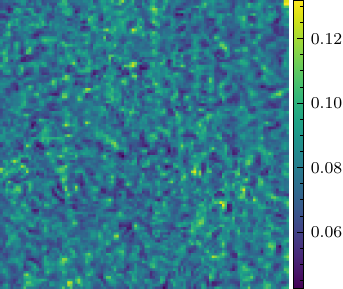}}
    \quad 

    \subfloat[$\z$ \label{fig:real_z_samsung}]{\includegraphics[height=\STDIMGSIZE\textwidth]{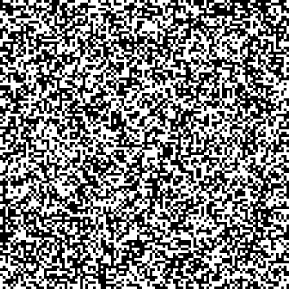}}
    \quad     
    \subfloat[Samsung $\xtot$ \label{fig:samsung x2t 2}]{\includegraphics[height=\STDIMGSIZE\textwidth]{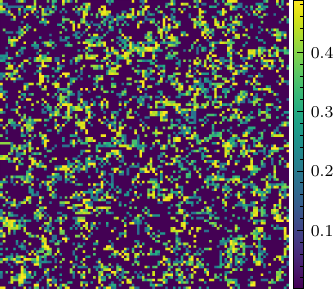}}
    \qquad \qquad
    \subfloat[Samsung $\x$ \label{fig:real_x_samsung}]{\includegraphics[height=\STDIMGSIZE\textwidth]{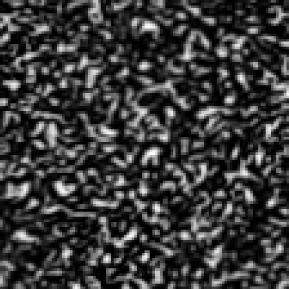}}
    \quad         
    \subfloat[Samsung $\ttox$ \label{fig:samsung t2x}]{\includegraphics[height=\STDIMGSIZE\textwidth]{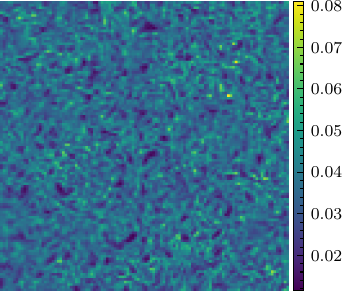}}
    \quad

    \subfloat[$\z$ \label{fig:real_z_scanner}]{\includegraphics[height=\STDIMGSIZE\textwidth]{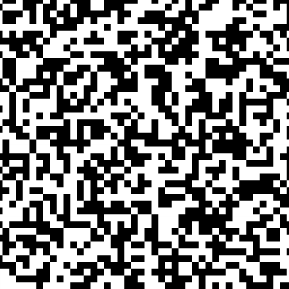}}
    \quad   
    \subfloat[Scanner $\xtot$ \label{fig:scanner x2t}]{\includegraphics[height=\STDIMGSIZE\textwidth]{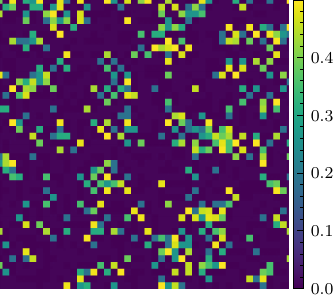}}
    \qquad \qquad
    \subfloat[Scanner $\x$ \label{fig:real_x_scanner}]{\includegraphics[height=\STDIMGSIZE\textwidth]{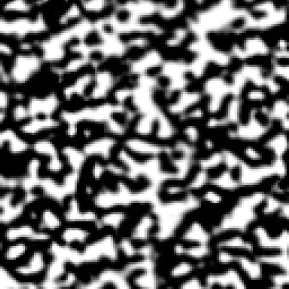}}
    \quad    
    \subfloat[Scanner $\ttox$ \label{fig:scanner t2x}]{\includegraphics[height=\STDIMGSIZE\textwidth]{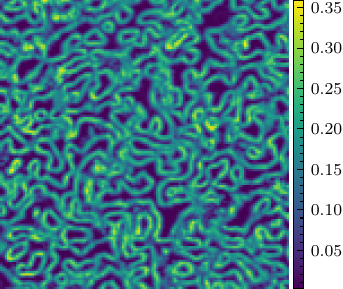}}
    } {
    \begin{tabular}{l l l c@{\hspace{30pt}} l l }
    & \multicolumn{1}{c}{$\z$} & \multicolumn{1}{c}{$\xtot$} & & \multicolumn{1}{c}{$\x$} & \multicolumn{1}{c}{$\ttox$} \\
    
    iPhone & \raisebox{-.5\height}{\includegraphics[height=\STDIMGSIZE\textwidth]{images/2d_std/gt_iphone_x2t.pdf}} & \raisebox{-.5\height}{\includegraphics[height=\STDIMGSIZE\textwidth]{images/2d_std/std_iphone_x2t.pdf}} & &\raisebox{-.5\height}{\includegraphics[height=\STDIMGSIZE\textwidth]{images/2d_std/gt_iphone_t2x.pdf}} & \raisebox{-.5\height}{\includegraphics[height=\STDIMGSIZE\textwidth]{images/2d_std/std_iphone_t2x.pdf}}  \\
    & & & & & \\
    Samsung & \raisebox{-.5\height}{\includegraphics[height=\STDIMGSIZE\textwidth]{images/2d_std/gt_samsung_x2t.pdf}} & \raisebox{-.5\height}{\includegraphics[height=\STDIMGSIZE\textwidth]{images/2d_std/std_samsung_x2t.pdf}} & &\raisebox{-.5\height}{\includegraphics[height=\STDIMGSIZE\textwidth]{images/2d_std/gt_samsung_t2x.pdf}} & \raisebox{-.5\height}{\includegraphics[height=\STDIMGSIZE\textwidth]{images/2d_std/std_samsung_t2x.pdf}} \\
    & & & & & \\
    Scanner & \raisebox{-.5\height}{\includegraphics[height=\STDIMGSIZE\textwidth]{images/2d_std/gt_scanner_x2t.pdf}} & \raisebox{-.5\height}{\includegraphics[height=\STDIMGSIZE\textwidth]{images/2d_std/std_scanner_x2t.pdf}} & &\raisebox{-.5\height}{\includegraphics[height=\STDIMGSIZE\textwidth]{images/2d_std/gt_scanner_t2x.pdf}} & \raisebox{-.5\height}{\includegraphics[height=\STDIMGSIZE\textwidth]{images/2d_std/std_scanner_t2x.pdf}}\\
    \end{tabular}
    }
    \caption{An example of 2D variability for a randomly selected CDP. We used pixel-wise standard deviation $\sigma$ to estimate the variability among the generated images. For better visual comprehension, we display a central crop that is equal to half the dimensions of the full image. %
    }
    \label{fig:2d_std}
    
\end{figure*}
}

\section{Related work}
\label{sec:related work}

\subsection{Turbo family}
\label{subsec:turbo}

The Turbo framework was derived based on the solid information theoretic foundations \cite{Belousov2022wifs}. That framework consists of two paths, i.e., direct and reverse ones. In the general case, both paths are trained simultaneously and share common training blocks. However, in particular cases, the framework might be trained in one path only. At the inference stage, both paths or just one of them might be used depending on the targeted application. As it was investigated in \cite{Belousov2022wifs}, Turbo can be trained on paired or unpaired data, providing flexibility in its application. Turbo generalizes pix2pix (paired)  \cite{pix2pix2017} and CycleGAN (unpaired) \cite{zhu2017unpaired} image-to-image translation systems. 

Turbo consists of several building blocks and losses that make the training procedure complex enough. %
Once trained, Turbo is quite efficient and very fast at the inference stage. The main drawback of Turbo is the deterministic nature of generated CDP, i.e., for the given input it provides only one output.

In \cite{Belousov2022wifs} we extensively investigated the impact of various factors such as the backbone architectures, the discriminator types, losses, etc., on the overall system's performance and found the optimal ones. In the current work, we use two found optimal configurations:
\begin{itemize}
    \item \TRBpairdOne
    \begin{displaymath}
      \begin{aligned}
          \mathcal{L}^{\text{paired (w } \mathcal{D} \text{)}}_{\text{CNN-RESNET-CNN}}(\phi, \theta)
           & = \Let + \Det    \\ 
           & + \lambda_D \Lrp  + \lambda_D \Drp           \\ 
           & + \lambda_T \Lep   + \lambda_T \Dep              \\ 
           & + \lambda_T \lambda_R \Lrt   + \lambda_T \lambda_R \Drt,
      \end{aligned}
    \end{displaymath}        
    \item \TRBpairdUNET
    \begin{displaymath}
      \begin{aligned}
          \mathcal{L}^{\text{paired (w/o } \mathcal{D} \text{)}}_{\text{UNET}}(\phi, \theta)
           & = \Let   + \lambda_D \Lrp \\ 
           & + \lambda_T \Lep  + \lambda_T \lambda_R \Lrt,
      \end{aligned}
    \end{displaymath}    
\end{itemize}
where $\x$ denotes the image of CDP, $\z$ denotes the digital template, $\xh$ and $\zh$ denote the reconstructions and $\xe$ and $\ze$ are the generated images. The terms $\Let$, $\Lrp$, $\Lep$ and $\Lrt$ are the conditional cross-entropy terms that are implemented as $\ell_1$-norm pair-wise losses between the corresponding entities. $\Det$, $\Drp$, $\Dep$ and $\Drt$ impose Kullback-Leibler (KL)-divergence constraints, i.e., the distribution matching losses, a.k.a. adversarial losses, between the corresponding distributions and the parameters $\lambda_{T}$, $\lambda_{D}$ and $\lambda_{R}$  trade-off the losses. The detailed development of Turbo's losses and the schematic representation of the direct and reverse paths are given in \cite{Belousov2022wifs} and the corresponding code.

\subsection{DDPM}
\label{subsec:ddpm}

The DDPM  \cite{ho2020denoising}  are based on the minimization of Fisher divergence, which is also closely linked with the KL-divergence, between the data distribution and the energy-based model approximating the data distribution. The core concept of DDPM is to use a score function representing the gradient of the logarithm of the energy-based model with respect to the data sample to suppress the dependence on the normalization constant, which is infeasible to compute in practice \cite{luo2022understanding}. Similar to Turbo, DDPM consists of forward and reverse paths. However, in contrast to Turbo, the DDPM forward path is not trainable and is based on the addition of noise to network input. 
The addition of noise with variable variance aims at ``interpolation'' of data distribution represented by sparse training data samples. The variable variance of noise should address the different regions of data distribution in a function of the estimated probability density function\cite{luo2022understanding}. 

From the point of view of the CDP nature, the trained Turbo model can produce the printing simulations and the digital template estimations simultaneously, while DDPM requires training of two separate models. With respect to the number of optimised losses, the DDPM training is simpler and includes only one loss. We use the \Palette model \cite{saharia2022palette} to implement conditional DDPM. For the $\ttox$ case, the model's loss is: 
\begin{displaymath}
    \begin{aligned}
    \mathcal{L}^{DDPM}(\varphi) = \mathbb{E}_{t, \z, \x, \boldsymbol{\epsilon}}\left[\left\|\boldsymbol{\epsilon}-g_\varphi\left(\sqrt{\bar{\alpha}_t} \x+\sqrt{1-\bar{\alpha}_t} \boldsymbol{\epsilon}, \textbf{z}, t\right)\right\|^2\right],
    \end{aligned}
    \label{eq:ddpm_simple_loss}
\end{displaymath}
where $\mathbf{x}$ denotes the target image, $\mathbf{z}$ denotes the digital template used as a conditioning, $\boldsymbol{\epsilon} \sim \mathcal{N}({\bf 0}, \mathbf{I})$ denotes the noise added at step $t$, $g_\varphi$ stands for the parametrized denoiser model, $\bar{\alpha}_t$ denotes the noise scale parameter \cite{ho2020denoising}. For the $\xtot$ channel modeling, a similar loss is used, but the digital template $\z$ is used as the target image, and the model is conditioned by $\x$. In contrast to Turbo, the DDPM loss does not allow the training on unpaired data. 

Training and inference stages are iterative and require adapting to many noise levels. Contrary to Turbo, which generates data in a single step, DDPM might need hundreds of steps to produce the final result. %
Despite this, a notable distinction between DDPM and Turbo lies in the stochastic nature of DDPM, enabling it to generate multiple outputs from a single input, thereby accommodating the intrinsic randomness associated with the printing process. The schematic block diagram of DDPM is shown in Fig. \ref{fig:ddpm scheme}.

\section{Dataset and training details}

\subsection{Dataset}

For empirical evaluation of the models under investigation we used the data acquired by two modern mobile phones and by a high-resolution scanner.

The experiments on the scanner data are an extension of our previous work \cite{Belousov2022wifs}. In this respect, the same Indigo $1 \times 1$ symbol dataset \cite{chaban2021wifs}\footnote{\url{http://sip.unige.ch/projects/snf-it-dis/datasets/indigo-base}} was used. This dataset consists of 720 digital templates of size $228 \times 228$ with $1 \times 1$ pixel symbol size. The digital templates have been printed at HP Indigo 7600 industrial printer at a resolution of 812.8 dpi and enrolled by Epson Perfection V850 Pro scanner at a resolution of 2400 ppi. Considering the ratio between the printing and acquisition resolutions, the obtained CDP are of size $684 \times 684$, i,e., $1 \times 1$ pixel in the digital template corresponds to a $3 \times 3$ block in the acquired CDP. The final codes are 16-bit grayscale images.

The experiments on the mobile phone data were performed on the recently created Indigo 1x1 variability dataset \cite{chaban2022wifs}\footnote{\url{http://sip.unige.ch/projects/snf-it-dis/datasets/indigo-variability}} that consists of 1440 digital templates of size $228 \times 228$ with $1 \times 1$ pixel symbol size. The templates have been printed at HP Indigo 5500 industrial printer at a resolution of 812.8 dpi and enrolled by iPhone 12 Pro and Samsung Galaxy Note 20 Ultra cell phones. The obtained CDP images are of size $228 \times 228$ and encoded as 8-bit RGB images. However, for the sake of simplicity, we convert them into grayscale images. 

Both mentioned datasets contain original and fake CDP. For our experiments, we used only the original codes. However, it should be noted that in both cases the fakes were produced on the same printing and acquisition equipment as the original codes. In this respect, the model trained on the original codes can be effectively applied to generate fake codes.

\begin{figure}[t!]
	\includegraphics[width=0.85\columnwidth]{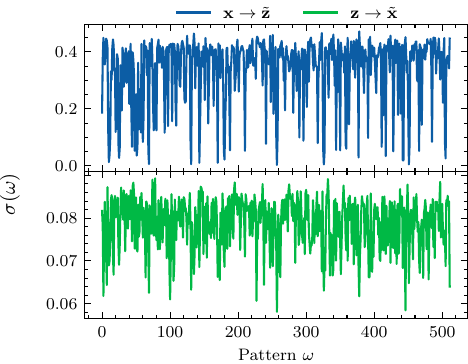}
    \caption{The x-axis represents the 512 different possible patterns $\omega$ ordered by their flattened binary representations. The y-axis represents the standard deviation of the central pixel of each pattern  for the iPhone. %
    }
    \label{fig:std_per_pattern_iphone} 
\end{figure}

\subsection{Training details}

The Turbo framework architectures' details and training conditions are the same as in \cite{Belousov2022wifs}. 

As a DDPM-based framework, we used the \Palette model \cite{saharia2022palette} with the UNET architecture inspired by \cite{dhariwal2021diffusion}. We modify UNET in the following way: it takes two channels as input, where the first channel is a noise and the second one is used for conditioning; the model incorporates attention with resolutions of 16 and includes two residual blocks per downsampling step. We initialized the model using Kaiming initialization. %
Additionally, we set the dropout rate to 0.2 to prevent overfitting due to the high similarity between the CDP. We train our models on a single A100 GPU with 80GB of memory with a mini-batch of size 36 for 15000 training epochs. 
We use a standard Adam optimizer with the $5e^{-5}$ learning rate and without a learning rate warmup schedule. Similarly to \cite{saharia2022palette}, we use 0.9999 EMA but, during the inference, we do not perform the hyper-parameter tuning over noise schedules and refinement steps. During training, we employ the same linear noise schedule of ($1e^{-6}$, 0.01) with 2000 time-steps and 1000 refinement
steps with a linear schedule of ($1e^{-4}$, 0.09) during inference as in \cite{saharia2022palette}
\footnote{
\new{The code and configuration files are publicly available at
\url{https://gitlab.unige.ch/sip-group/stochastic-digital-twin}}
}.

\section{Results and discussion}

\subsection{\Palette model stochasticity}

Fig. \ref{fig:stochasticity} demonstrates several examples of the diverse outputs of the \Palette model produced for the same randomly selected input from the iPhone subset. The top row corresponds to the modeling of printing channel $\ttox$ and the bottom one shows the digital template estimation channel $\xtot$. %

\begin{figure}[t!]
	\includegraphics[width=0.85\columnwidth]{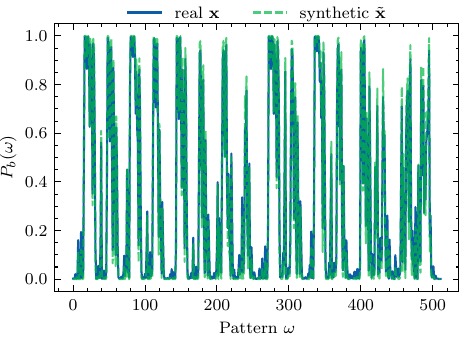}
     \caption{The x-axis represents the 512 different possible patterns $\omega$ ordered by their flattened binary representations. The y-axis represents the probability of bit-flipping for the central pixel of each pattern computed from iPhone data.} 
    \label{fig:bit_flipping_iphone}
\end{figure}
\newcommand{\MEANIMGSIZE}{0.685}
\begin{figure*}[t!]
    \centering
    \subfloat[Hamming distance\label{hamming}]{\includegraphics[width=\MEANIMGSIZE\columnwidth]{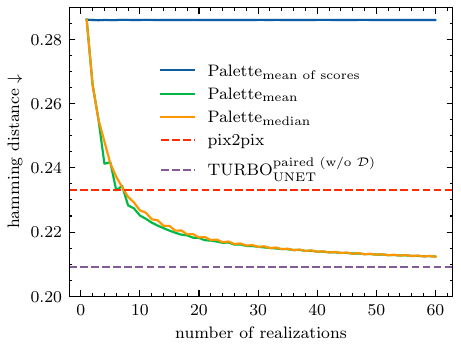}}
    \subfloat[MSE\label{mse}]{\includegraphics[width=\MEANIMGSIZE\columnwidth]{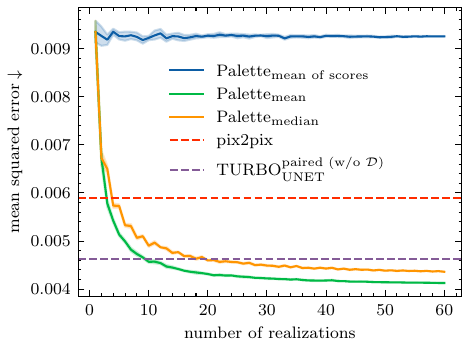}}
    \subfloat[SSIM\label{ssim}]{\includegraphics[width=\MEANIMGSIZE\columnwidth]{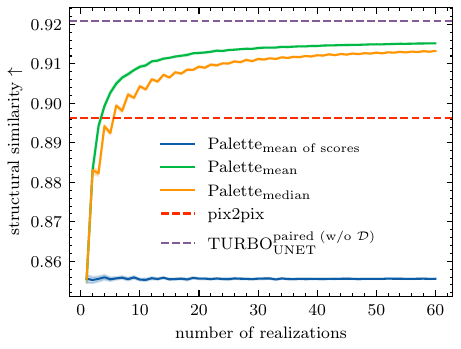}}
    \caption{Impact of the number of realizations on different metrics for the iPhone dataset.}
    \label{fig:mean_score_iphone}
\end{figure*}

To illustrate the variability in the generated data, we picked the same template and stacked the produced outputs as a 3D tensor, then we calculated the standard deviation in the image dimension, i.e., for each pixel of generated images. The obtained results are visualized in Fig. \ref{fig:2d_std}. 
The left part of the figure shows the results for the $\xtot$ %
for the iPhone, Samsung, and scanner, respectively. It is straightforward to observe numerous regions characterized by diminutive standard deviation, as indicated by the dark blue hue. These regions symbolize the model's degree of confidence in the generated outcomes, which correlates to the conglomerations of white and black pixels. Conversely, the yellow hue denotes areas of heightened standard deviation, reflective of the model's uncertainty. These areas typically align with the transitional regions, manifesting the boundary conditions between different pixel clusters. It is important to note that these features are best seen in the case of the scanner due to the higher acquisition resolution. 
The right part %
shows the results for the $\ttox$ channel. %
In contrast to the $\xtot$ channel, the general dynamic range for the obtained standard deviation is about \numrange{3}{5} times smaller. This can be explained by the fact that the printed images are more continuous, i.e., have a more uniformly distributed histogram, which makes the image synthesis more reliable. For Samsung, the dynamic range of the obtained deviation is about 1.5 times smaller than for iPhone. For the scanner results one also observes a smaller amount of unreliable regions with the edge-transition regions being very well pronounced.

\begin{table}[t!]
\renewcommand{\arraystretch}{\AS}
\centering
\caption{Models performance on the iPhone data}
\label{table:final_iphone}
\begin{tabular}{lccccc}
    \toprule
    \scriptsize 
    \multirow{2}{*}{\textbf{\scriptsize Model}} & 
    \textbf{\scriptsize FID} & 
    \textbf{\scriptsize Hamm.} & 
    \textbf{\scriptsize FID} & 
    \multirow{2}{*}{\textbf{\scriptsize MSE}} & 
    \multirow{2}{*}{\textbf{\scriptsize SSIM}} \\
    & 
    \textbf{\scriptsize $\x$\textrightarrow $\ze$} & 
    \textbf{\scriptsize dist.} & 
    \textbf{\scriptsize $\z$\textrightarrow $\xe$} & &  \\
    \midrule
    
    {\IT  W/O processing} & \IT 289.68 & \IT 0.300 & \IT 289.68 & \IT 0.254 & \IT 0.249 \\ 
    {\pixp} & 11.82 & 0.232 & 11.64 & 0.005  & 0.910 \\
    {\CyG} & 20.69 & 0.268  & 12.59 & 0.014  & 0.782 \\
    {\TRBpairdOne} & 6.56 & 0.239 & 10.20 & 0.005 & 0.915 \\
    {\TRBpairdUNET } & 35.35 & \B 0.210 & 12.22 & \B 0.004 & \B 0.925 \\
    {$\textrm{Palette}_\textrm{mean}$} & \B 4.64 & 0.211 & \B 9.00 & \B 0.004 & 0.915 \\

    \bottomrule
\end{tabular}
\end{table}

It was shown in \cite{Tutt2022wifs} that the printed pixel's variability depends on the surrounding neighborhood that we refer to as \textit{pattern} $\omega$, where $\omega$ denotes a $3 \times 3$ configuration of each pattern. To investigate if the same effect is present in the synthetically generated codes we define $2^{(3 \times 3)} = 512$ possible patterns %
and calculate the standard deviation of the central pixel for each of them through all generated codes. The results obtained for the iPhone dataset are shown in Fig. \ref{fig:std_per_pattern_iphone}. For Samsung, we observed quite a similar picture. In Fig. \ref{fig:std_per_pattern_iphone} we can see the same tendency as in Fig. \ref{fig:2d_std}, namely, the general dynamic range of the standard deviation for the $\xtot$ channel is higher than for the $\ttox$, i.e., \numrange{0}{0.45} versus \numrange{0.06}{0.09}. Also, we observe the pattern dependence for both channels but this dependence differs between the channels that is natural. In particular, in the $\xtot$ channel there are more patterns with the less variable central pixel, i.e., $\sigma(\omega)$ close to 0. 

To study the similarity between the real CDP $\x$ and the synthetic counterparts $\xe$, we compute the probability of bit-flipping for the central pixel for each pattern after Otsu binarization, as suggested in \cite{Tutt2022wifs}. The results obtained for the iPhone are shown in Fig. \ref{fig:bit_flipping_iphone}\footnote{For Samsung we observed the same tendency.}. We can see that some patterns almost certainly flip with $P_b(\omega)$ close to 1, whereas others produce reliable results with $P_b(\omega)$ close to 0. But the most important thing is that the bit-flipping probability for the real $\x$ perfectly correlates with one for the synthetic $\xe$. 

\begin{table}[t!]
\renewcommand{\arraystretch}{\AS}
\centering
\caption{ Models performance on the Samsung data}
\label{table:final_samsung}
\begin{tabular}{lccccc}
    \toprule
    \scriptsize 
    \multirow{2}{*}{\textbf{\scriptsize Model}} & 
    \textbf{\scriptsize FID} & 
    \textbf{\scriptsize Hamm.} & 
    \textbf{\scriptsize FID} & 
    \multirow{2}{*}{\textbf{\scriptsize MSE}} & 
    \multirow{2}{*}{\textbf{\scriptsize SSIM}} \\
    & 
    \textbf{\scriptsize $\x$\textrightarrow $\ze$} & 
    \textbf{\scriptsize dist.} & 
    \textbf{\scriptsize $\z$\textrightarrow $\xe$} & &  \\
    \midrule
  
    \IT W/O processing  & \IT 381.44 & \IT 0.314 & \IT 381.44 & \IT 0.278 & \IT 0.193 \\
    \pixp & 8.53 & 0.241 & 20.18 & \B 0.004 & 0.908 \\
    \CyG  & 8.85 & 0.283 & 22.85 & 0.015 & 0.694 \\
    \TRBpairdOne & 7.01 & 0.247 & 17.34 & \B 0.004 & 0.914 \\
    \TRBpairdUNET & 54.80 & \B 0.211 & 28.88 & \B 0.004 & \B 0.922 \\
    $\textrm{Palette}_\textrm{mean}$ & \B 4.46 & 0.215 & \B 10.72 & \B 0.004 & 0.908\\
    \bottomrule
\end{tabular}
\end{table}

\subsection{Aggregation techniques}

For the \Palette model, we study the impact of the number of realizations and different aggregation techniques. The iPhone results are shown in Fig. \ref{fig:mean_score_iphone}\footnote{The results for the Samsung data are similar to iPhone.}, where $\textrm{Palette}_\textrm{mean}$ denotes that the final output is obtained as a mean of predictions, while in $\textrm{Palette}_\textrm{median}$ we take a median of predictions, and then the     score is calculated for the aggregated prediction. In the case $\textrm{Palette}_\textrm{mean of scores}$ the reference metric is calculated for each prediction and then the mean value of obtained scores is taken.  

One can observe that $\textrm{Palette}_\textrm{mean of scores}$ error is almost constant, %
while the errors for $\textrm{Palette}_\textrm{mean}$ and $\textrm{Palette}_\textrm{median}$ are close to each other and decrease as the number of realizations increases.
This can be explained by the fact that from one side the global error is the same at each realization but the local errors appear in different positions. The aggregation allows us to reduce them but, at the same time, it leads to the loss of stochasticity. In the same plot, one can see the results for pix2pix and \TRBpairdUNET models, but these models are deterministic, and their results are not impacted by the number of realizations. \Palette easily outperforms pix2pix on all metrics after 5-7 reali\new{z}ations. In terms of Hamming distance and SSIM \Palette is not capable of outperforming \TRBpairdUNET.
For MSE, \Palette needs at least \numrange{10}{20} realizations to surpass Turbo.

\subsection{Models general performance}

{
\renewcommand{\arraystretch}{\AS}

\begin{table}
\centering
\caption{Models performance on the Scanner data}
\label{table:final_scanner_2021}
\begin{tabular}{lccccc}
    \toprule
    \scriptsize 
    \multirow{2}{*}{\textbf{\scriptsize Model}} & 
    \textbf{\scriptsize FID} & 
    \textbf{\scriptsize Hamm.} & 
    \textbf{\scriptsize FID} & 
    \multirow{2}{*}{\textbf{\scriptsize MSE}} & 
    \multirow{2}{*}{\textbf{\scriptsize SSIM}} \\
    & 
    \textbf{\scriptsize $\x$\textrightarrow $\ze$} & 
    \textbf{\scriptsize dist.} & 
    \textbf{\scriptsize $\z$\textrightarrow $\xe$} & &  \\
    \midrule

    \IT W/O processing  & \IT 304.13 & \IT 0.238 & \IT 304.01 & \IT 0.181 & \IT 0.480 \\
    \pixp & 3.37 & 0.111 & 8.57 & 0.045 & 0.758 \\
    \CyG  & 3.87 & 0.155 & 4.45 & 0.049 & 0.732 \\
    \TRBpairdOne & 3.16 & 0.086 & 6.60 & 0.040 & \B 0.779 \\
    \TRBpairdUNET  & 6.21 & 0.100 & 28.110 & \B 0.036 & 0.778 \\
    $\textrm{Palette}_\textrm{mean}$ & \B 2.90 & \B 0.081 & \B 4.36 & 0.038 & 0.769  \\
    \bottomrule
\end{tabular}
\end{table}

}

For further performance evaluation, we use $\textrm{Palette}_\textrm{mean}$. 
The inference time for a single realization of the \Palette model applied to 280 test images is approximately 30 minutes. In contrast, the Turbo model achieves an inference time of just 15 seconds on the same GPU for the same dataset. Considering the inference time complexity, we found that using 21 realizations strikes a good balance between reasonable inference execution time, stochasticity preservation, and the accuracy of the final result.
We compare the performance of the \Palette in this configuration with the state-of-the-art pip2pix \cite{pix2pix2017}, CycleGAN \cite{zhu2017unpaired}, \TRBpairdOne, and \TRBpairdUNET%
\new{on the same set of metrics as in \cite{Belousov2022wifs}.}
\textit{W/O processing} setup is used to estimate the baseline performance where we assume $\ze = \x$ and $\xe = \z$, i.e., an ideal printing-imaging channel without any distortions. 
The results obtained for the data enrolled by the iPhone and Samsung mobile phones are given in Tables \ref{table:final_iphone} and \ref{table:final_samsung} respectively. 

It should be noted that the results obtained for the FID metric for the $\xtot$ and $\ttox$ channels are quite unstable and differ a lot between the models. This can be explained by the fact that FID was developed %
for natural images while, in our case, all CDP %
significantly differ from them. %
The other metrics demonstrate more coherent results. %

For both mobile phones, the MSE results are almost identical for \Palette and both Turbo configurations. In terms of SSIM, the results are also very close with a slight \TRBpairdUNET superiority. \TRBpairdUNET  also outperforms the other models in terms of Hamming distance. The results for the remaining models exhibit marginal inferiority. 

The result obtained for the data enrolled by the scanner are given in Table \ref{table:final_scanner_2021} and the general tendency is the same, namely, FID behavior is very unstable; CycleGAN demonstrates the worst results; the results for the \Palette and Turbo are quite close but Palette is slightly superior to Turbo on the Hamming distance, albeit at the cost of significantly higher complexity of inference.

\section{Conclusion}

The current work is a continuation of our previous study \cite{Belousov2022wifs} related to modeling of complex physical printing-imaging processes using a machine learning based models known as a \textit{digital twin} for anti-counterfeiting applications based on CDP. The current work is dedicated to investigation of the applicability of DDPM for such modeling. 

Our main interest was to explore the stochasticity of DDPM. The obtained results show that synthetic digital and CDP images are close enough to the real ones in terms of considered metrics. Moreover, the synthetic CDP images produced by DDPM fully reflect the natural randomness of the printing process. This makes the DDPM-based model a suitable candidate for the role of a synthetic generator. The general performance of the studied \Palette model is comparable to that of the Turbo framework. The main drawback of the DD\new{PM} is the computation complexity of the inference stage. 

The investigation of more advanced sampling techniques at the training and inference stages for the improvement of DD\new{PM} complexity is the main direction for our future work.

\bibliographystyle{IEEEtran}
\bibliography{bibliography}

\end{document}